\documentclass[10pt,twocolumn,letterpaper]{article}

\usepackage[pagenumbers]{wacv} 

\usepackage[accsupp]{axessibility}  

\usepackage{graphicx}
\usepackage{amsmath}
\usepackage{amssymb}
\usepackage{booktabs}

\usepackage{enumitem}
\usepackage[normalem]{ulem}
\usepackage{array}
\usepackage{textcomp}
\usepackage{mathtools}
\usepackage[flushmargin]{footmisc}
\usepackage{colortbl}
\usepackage{booktabs}
\usepackage{tabularx}
\usepackage{ctable}
\usepackage{pifont}
\usepackage{multirow}

\usepackage[pagebackref,breaklinks,colorlinks]{hyperref}

\newcommand{\be}{\begin{eqnarray}}
\newcommand{\ee}{\end{eqnarray}}
\newcommand{\bee}{\begin{eqnarray*}}
\newcommand{\eee}{\end{eqnarray*}}

\newcommand{\matrixb}{\left[ \begin{array}}
\newcommand{\matrixe}{\end{array} \right]}

\newcommand{\app}{\raise.17ex\hbox{$\scriptstyle\sim$}}
\input{def_custom.tex}

\newcommand{\newpara}[1]{\vspace{6pt}\noindent\textbf{#1}}
\setlist[itemize]{align=parleft,left=0pt,topsep=1mm,itemsep=0mm,parsep=1mm}
\def\ie{\textit{i.e}\onedot}

\usepackage[capitalize]{cleveref}
\crefname{section}{Sec.}{Secs.}
\Crefname{section}{Section}{Sections}
\Crefname{table}{Table}{Tables}
\crefname{table}{Tab.}{Tabs.}

\def\wacvPaperID{2039} 
\def\confName{WACV}
\def\confYear{2024}

\begin{document}

\title{Can CLIP Help Sound Source Localization?}

\author{Sooyoung Park$^*$$^{1,2}$ \quad
Arda Senocak$^*$$^{1}$ \quad
Joon Son Chung$^{1}$ \\
$^1$ Korea Advanced Institute of Science and Technology,  South Korea \\
$^2$ Electronics and Telecommunications Research Institute, South Korea
}

\maketitle

{\let\thefootnote\relax\footnote{$^*$These authors contributed equally to this work. This work was supported by Electronics and Telecommunications Research Institute (ETRI) grant funded by the Korean government, [23ZH1200, The research of the basic media$\cdot$contents technologies] and the MSIT (Ministry of Science and ICT), Korea, under the ITRC (Information Technology Research Center) support program (IITP-2023-00259991) supervised by the IITP (Institute for Information \& Communications Technology Planning \& Evaluation).}}

\begin{abstract}
Large-scale pre-trained image-text models demonstrate remarkable versatility across diverse tasks, benefiting from their robust representational capabilities and effective multimodal alignment. We extend the application of these models, specifically CLIP, to the domain of sound source localization. Unlike conventional approaches, we employ the pre-trained CLIP model without explicit text input, relying solely on the audio-visual correspondence. To this end, we introduce a framework that translates audio signals into tokens compatible with CLIP's text encoder, yielding audio-driven embeddings. By directly using these embeddings, our method generates audio-grounded masks for the provided audio, extracts audio-grounded image features from the highlighted regions, and aligns them with the audio-driven embeddings using the audio-visual correspondence objective. Our findings suggest that utilizing pre-trained image-text models enable our model to generate more complete and compact localization maps for the sounding objects. Extensive experiments show that our method outperforms state-of-the-art approaches by a significant margin.
\end{abstract}
\vspace{-8mm}
\section{Introduction}\label{sec:intro}
The ability of humans and other animals to pinpoint the locations of sound sources is crucial for perceiving the world around us. We receive continuous multisensory information, such as auditory and visual inputs, understand their relationships, infer which object/event is producing sound, and focus on sounding objects/events. To provide machine perception with similar abilities, audio-visual sound source localization has been extensively explored in recent years~\cite{senocak2018learning,arandjelovic2018objects,senocak2019learning,qian2020multiple,hu2020discriminative,chen2021localizing,lin2021unsupervised,li2021space,senocakLessMore,song2022sspl,senocakHardPos,ssslTransformation,ezvsl,marginnce,slavc,sun2023learning,senocak2023alignment}. One fundamental approach in this direction involves leveraging the natural correspondence between audio and visual signals without explicit supervision or the need for annotated data. The most predominant method for achieving this is by aligning audio-visual representations as a self-supervision signal within a contrastive learning framework.

While sound localization methods are trained with the aforementioned fundamental assumption, some additional prior knowledge is also incorporated. These pieces of prior knowledge are introduced in the form of using visual objectness~\cite{ezvsl,slavc} and object proposal networks~\cite{xuan2022proposal}, or other modalities such as optical flow~\cite{htf}. As true sound source localization methods necessitate a strong audio-visual semantic alignment, the previously mentioned priors might not contribute to improved alignment, as they can introduce visual objectness or motion bias that may lead to shortcuts~\cite{oya2020we,slavc,arandjelovic2018objects}. In this work, our focus is to leverage strong multimodal alignment knowledge as a prior to improve audio-visual alignment for genuine sound source localization. From this perspective, we employ the Contrastive Language-Image Pretraining (CLIP)~\cite{radford2021learning} model for the sound source localization task. This choice is due to its robust representation and multimodal alignment capability, stemming from learning directly from raw text about images on large scale data. Thus, it provides a broader source of supervision rather than limited category labels.

\begin{figure}[t]
\centering
        \includegraphics[width=1.0\linewidth]{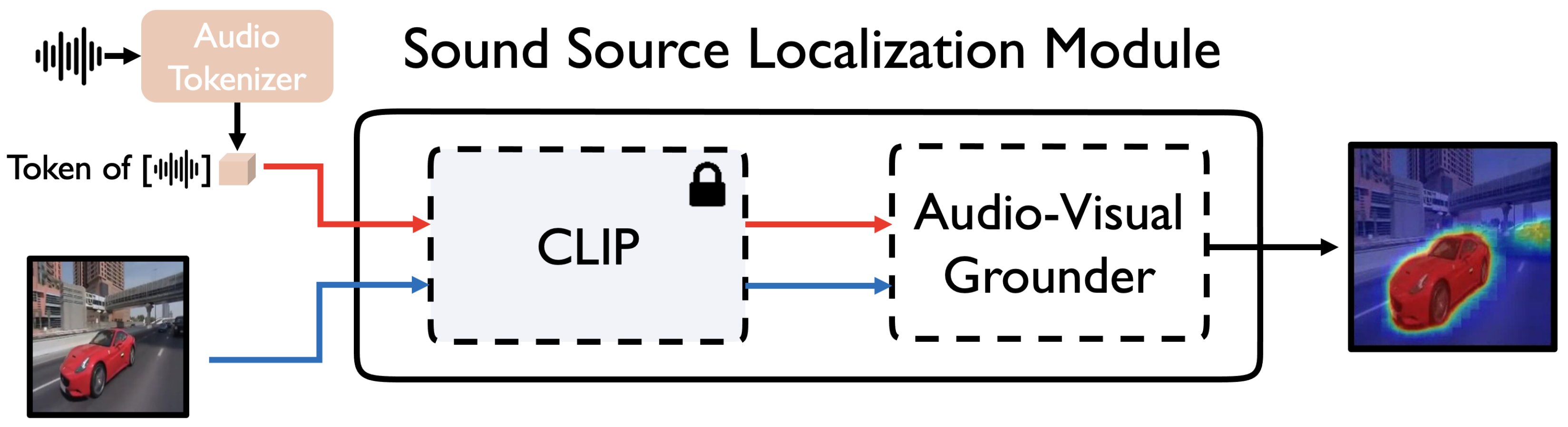}
\caption{\textbf{The proposed text input-free CLIP based sound source localization method.} 
}
\label{fig:teaser}
\vspace{-6mm}
\end{figure}

The frameworks that leverage the CLIP model generally include text queries/prompts. However, we aim to explore this approach without using explicit contextual text information. The reasons we do not intuitively utilize direct text inputs are as follows: (1) There is no available paired text data in sound source localization benchmark datasets, (2) the sound source localization task is unlabeled, (3) a genuine sound source localization approach necessitates learning pure audio-visual alignment through self-supervision. Therefore, in this paper, we employ the pre-trained CLIP model in a textless manner (as illustrated in~\Fref{fig:teaser}), relying solely on audio-visual correspondence.

\begin{figure*}[tp]
    \centering
    \includegraphics[width=0.95\linewidth]{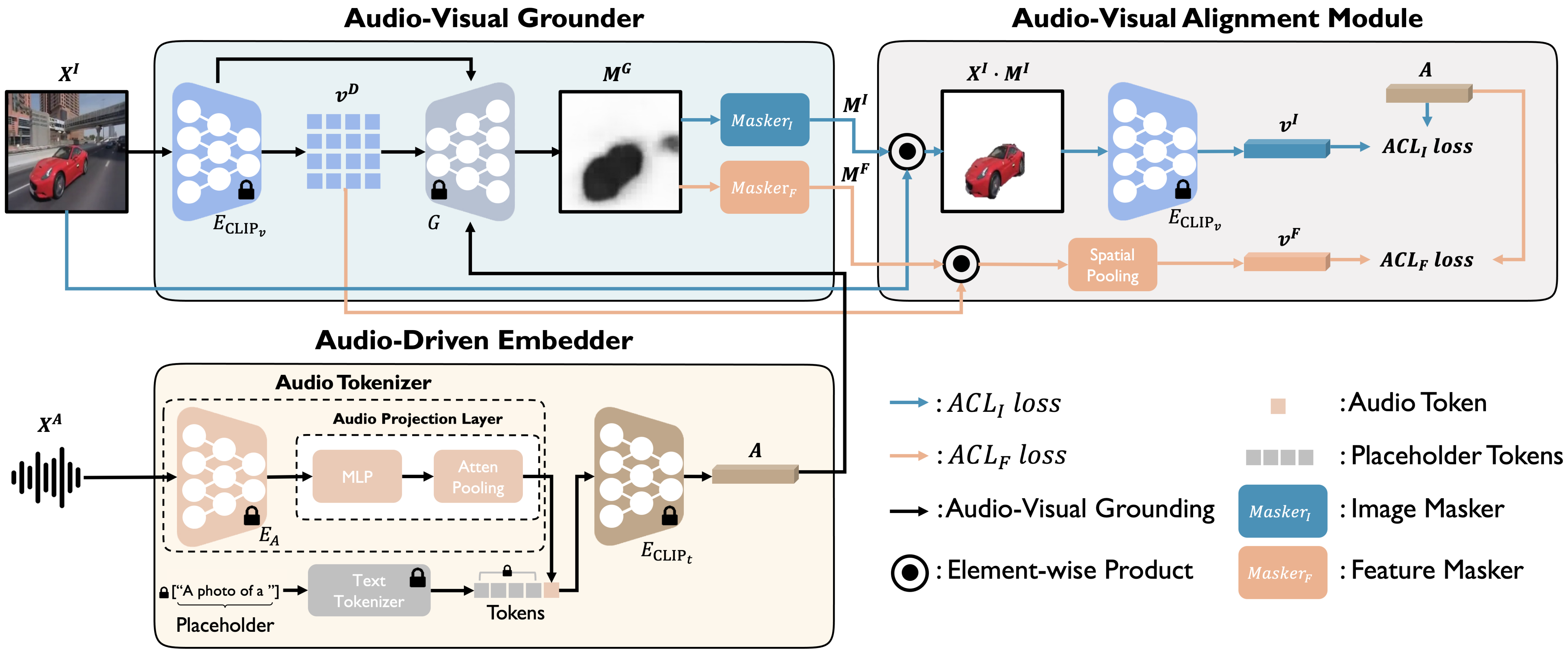}
    \caption{{\bf Our sound source localization framework.} The proposed method takes audio-visual pairs, translating audio signals into CLIP-compatible tokens via the Audio Tokenizer module to generate audio-driven embedding, $\mathbf{A}$. This embedding highlights sounding regions within the Audio-Visual Grounder module. With the sounding area masks, the Audio-Visual Alignment module extracts audio-grounded visual features at both image-level ($\boldsymbol{v}^I$) and feature-level ($\boldsymbol{v}^F$). These visual features and audio feature are aligned via contrastive learning.}
    \label{fig:pipeline}
    \vspace{-4mm}
\end{figure*}

To utilize CLIP in a text input-free manner and train our sound source localization method through self-supervised audio-visual alignment, we propose the following steps (depicted in~\Fref{fig:pipeline}): First, we introduce a framework that translates audio signals into tokens compatible with CLIP's text encoder. This process yields contextual embeddings for the provided audio input, a concept we refer to as audio-driven embedding. Second, our key idea involves aligning audio and visual features in a self-supervised manner using contrastive learning. Consequently, we seamlessly integrate this audio-driven embedding to emphasize the sounding regions within the visual scenes. Subsequently, audio-grounded visual features on both the image and feature levels are extracted from these regions. These features are then aligned with the audio-driven embedding through audio-visual correspondence within a contrastive learning framework. The entire model is trained at once with the audio-visual alignment objective. Through our experiments, we validate that the proposed method outperforms existing approaches and baselines. In some instances, it even achieves competitive results when compared to fully supervised or text-queried sound source localization baselines.

We summarize the contributions of our work as follows:
\begin{itemize}
    \item We present a novel self-supervised sound source localization framework that exploits the large-scale pre-trained CLIP model.
    \item We propose an end-to-end textless approach, \ie no explicit text input. Our framework translates audio signals into tokens that are compatible with CLIP to obtain audio-driven embeddings.
    \item We utilize the audio-driven embeddings to emphasize the sounding regions and align them with the audio content for the objective of audio-visual correspondence.
    \item We conduct extensive experiments on the VGG-SS, SoundNet-Flickr, VGG-SS OpenSet, AVSBench, and Extended VGG-SS/SoundNet-Flickr datasets, collectively demonstrate the effectiveness of our proposed method.
\end{itemize}

\vspace{-2mm}
\section{Related work}\label{sec:RW}
\vspace{-2mm}
\newpara{Sound source localization.} The predominant technique employed for audio-visual sound source localization involves cross-modal attention~\cite{senocak2018learning,tian2018audio,senocak2019learning}, often coupled with contrastive loss. Following the contrastive learning paradigm, subsequent enhancements have been made by explicitly incorporating hard negatives from background regions~\cite{chen2021localizing}, utilizing iterative contrastive learning with pseudo-labels obtained from the same model in previous epochs~\cite{lin2021unsupervised}, applying transformation invariance and equivariance through data augmentations and geometric consistency~\cite{ssslTransformation}, considering semantically similar hard positives~\cite{senocakHardPos}, implementing negative-free contrastive learning~\cite{song2022sspl} similar to SiamSiam~\cite{chen2021exploring}, using momentum encoders to mitigate overfitting~\cite{slavc}, adding negative margin into contrastive learning alleviate the effect of noisy correspondences~\cite{marginnce}, and applying false negative-aware contrastive learning via intra-modal similarities~\cite{sun2023learning}. Following a similar trend, our method also integrates self-supervised contrastive learning.

Besides this trend, some other sound localization methods attempt to utilize additional prior knowledge or post-processing approaches.~\cite{senocakLessMore, qian2020multiple} incorporate label information to learn backbone audio and visual networks or to refine the audio-visual alignment. Xuan \etal~\cite{xuan2022proposal} use object priors in the form of object proposals, while Mo \etal~\cite{ezvsl} employ a post-processing approach to refine audio-visual localization results using pre-trained visual feature activation maps. In our work, we leverage CLIP's multimodal alignment knowledge as a prior in a textless and fully self-supervised manner without any post-processing.

\newpara{CLIP in Audio-Visual Learning.} Recent contrastive language-image pretraining (CLIP) models, which are pretrained on large-scale paired data~\cite{radford2021learning,jia2021scaling}, demonstrate robust generalization ability and have been successfully used in numerous downstream tasks across various research topics. In this section, we review related works that incorporate CLIP~\cite{radford2021learning} for audio-visual learning. WAV2CLIP~\cite{wu2022wav2clip} and AudioCLIP~\cite{guzhov2022audioclip} expand the pre-trained CLIP model by aligning audio features with text and visual features in a shared embedding space, \ie representation learning. They achieve this either using paired data or by utilizing the visual modality as a bridge.
Beyond representation learning, CLIP models are also employed in audio-visual event localization~\cite{mahmud2023ave} and video parsing~\cite{fan2023revisit}, as well as audio-visual source separation~\cite{tan2023language,dong2022clipsep}. While~\cite{tan2023language} employs text input for separation, CLIPSep~\cite{dong2022clipsep} is trained based on the audio-visual relationship without text query. Similarly, our proposed method is also trained solely with an audio-visual alignment objective. Another line of work~\cite{yariv2023audiotoken,bhatisegmental} adapt pre-trained CLIP models and text encoders for audio. They achieve this by mimicking contextual text tokens using audio signals, enabling the CLIP text encoder to embed audio signals. Our work also employs a similar approach to leverage the CLIP model without text input for the sound localization task.

\vspace{-2mm}
\section{Method}\label{sec:MTD}
\subsection{Audio-Driven Embedder} \label{ssec:tokenizer}
Our goal is to use the CLIP text encoder to embed audios without any text input. We employ the Audio Tokenizer module for this purpose, which transforms audio context into text-like tokens. In essence, an audio segment is translated into a word token, which can subsequently undergo processing by the pre-trained CLIP text encoder. The module has two key components: an \textbf{audio encoder} and a \textbf{projection network}. The projection network contains two MLP layers and one attentive pooling layer, similar to~\cite{yariv2023audiotoken}. While the audio encoder is pre-trained and fixed during training, the remaining layers are trained end-to-end in our sound source localization approach with the objective of audio-visual alignment.

\textbf{Audio Encoder}, $E_{A}$, is a transformer-based network pre-trained in a self-supervised manner, following~\cite{chen2022beats}. It takes an audio spectrogram and extracts audio embeddings. Once the audio embedding is extracted, it undergoes processing in a small \textbf{projection network}, effectively mimicking the textual tokens through the audio. The outcome is an ``audio token'' aligned with textual tokens. This token is then appended to the fixed placeholder text tokens of “A photo of a” to complete the input token representation as in~\Fref{fig:pipeline}. This way, the audio signal with its proper context can be fitted as an additional token for CLIP text encoder ($E_{\text{CLIP}_t}$). This combination of fixed placeholder text and audio tokens is processed in the pre-trained $E_{\text{CLIP}_t}$, and the audio-driven embedding $\mathbf{A}$ is obtained. This audio-driven embedding can then be paired or conditioned with any CLIP image encoder-based approaches as it contains the visual alignment knowledge due to $E_{\text{CLIP}_t}$.

\subsection{Audio-Visual Grounder} \label{ssec:grounder}
For a given input batch of audio-visual pairs, which consist of images and their corresponding audios, our audio-visual grounder performs grounding to detect the regions with sound and then generates masks. These masks are subsequently utilized to extract visual embeddings at both the image-level and feature-level, which are used in the audio-visual alignment objective. Our Audio-Visual grounding module is designed with three components: 1) an image encoder, 2) a grounder, and 3) mask generators.

We use a pre-trained CLIP image encoder as our image encoder, denoted as $E_{\text{CLIP}_v}$. It is responsible for encoding the provided input images into both global features and spatial features. For our grounder, $G$, we employ off-the-shelf CLIP-based segmentation network known as CLIPSeg~\cite{luddecke2022image}. It is important to note that CLIPSeg requires CLIP-based visual features and text conditioning to perform segmentation. We leverage the outputs from our image encoder as visual features for grounder. However, since our approach does not use any text input directly, we utilize our audio-driven embedding, $\mathbf{A}$, for conditioning. The result of the grounder $G$, $\mathbf{M}^G$, is potential sounding regions. Both the image encoder and the grounder remain fixed during training.

To obtain audio-grounded visual embeddings for the provided paired images $\mathbf{X}^I$ and audios $\mathbf{X}^A$ within the Audio-Visual Alignment module during training, it is essential to have differentiable binary masks for sounding regions. We introduce two masking methods: Image Masker ($Masker_I$) and Feature Masker ($Masker_F$), both of which serve to extract audio-grounded visual embeddings at the image-level and feature-level, respectively. Similar to~\cite{cha2023learning}, $Masker_I$ utilizes a learnable scalar projection $(w\cdot \mathbf{M}^G+b)$ on the output of the grounder, $\mathbf{M}^G$, and then applies the Gumbel-Max technique~\cite{jang2016categorical} to generate a differentiable binary mask, referred to as $\mathbf{M}^I$. This mask is used to identify sounding areas in the image. $Masker_F$ is designed with min-max normalization and soft-thresholding functions applied to $\mathbf{M}^G$ to obtain $\mathbf{M}^F$ , which allows the extraction of audio-visually correlated areas at the feature level. The utilization of these maskers is explained in the following section.

\vspace{-2mm}
\subsection{Audio-Visual Alignment} \label{ssec:alignment}
After obtaining sounding area masks for the given audio from the audio-visual grounder, our method extracts visual embeddings from the masked areas at both the image-level and feature-level, aligning them with the audio-driven embedding, $\mathbf{A}$, for the audio-visual alignment objective. For this purpose, we define two contrastive learning losses: image-level and feature-level audio-grounded contrastive losses, $ACL_I$ and $ACL_F$ respectively. In a nutshell, our model learns to maximize the alignment between the visual features of sounding regions and audio features.

\newpara{Image-Level Audio-Grounded Contrastive Loss.} Different from typical global image and audio correspondence, our focus is on alignment between sounding region and audio. One approach to achieve this is by highlighting the sounding regions (foreground pixels) in the image and masking out the background areas, as depicted in~\Fref{fig:pipeline}. To begin, the mask $\textbf{M}_i^I$ obtained from $Masker_I$ for audio-visual pair of $i$th clip to mask out the irrelevant areas in the image. Our image-level audio-grounded contrastive loss, $ACL_I$, consists of CLIP image encoder $E_{\text{CLIP}_v}$. This masked image is then transformed into a visual embedding, ${\boldsymbol{v}}^I_i= E_{\text{CLIP}_v}\left( \textbf{M}_i^I\cdot \textbf{X}^I_i \right)$. The audio-visual similarity between the audio-driven embedding $\mathbf{A}_j$ from $j$th clip and the audio-grounded visual embedding $\boldsymbol{v}_i^I$ is computed using cosine similarity and defined as $S^I_{i,j} = ({{\boldsymbol v}_i^I}\cdot \mathbf A_j)$. We employ symmetric InfoNCE for the contrastive loss. We note that image-level masks are computed only for positive pairs. Thus, the objective of this loss is to maximize the similarity between the positive sounding region and the corresponding audio pair, while also ensuring dissimilarity between negative audios and the actual sounding region. The $ACL_I$ loss is defined as follows:

\begin{align}
    \small
    \mathcal L_{ACL_I} = & \; \textit{InfoNCE}(\textbf{S}^I) \nonumber \\
    = &-\frac1{2B} \sum^B_i \log \frac{\exp(S^I_{i,i}/\tau)}{\sum^B_j \exp(S^I_{i,j}/\tau)} \nonumber \\
    & - \frac1{2B} \sum^B_i \log \frac{\exp(S^I_{i,i}/\tau)}{\sum^B_j \exp(S^I_{j,i}/\tau)}
    \label{eq:acli}
\end{align}

\noindent where $\tau$ is the temperature parameter and $\textbf{S}^I$ is image-level audio-visual similarity matrix within batch. With the help of this loss, the sounding region and the generated mask $\textbf{M}^I$ gradually cover the target sounding area. However, we observe that $ACL_I$ alone can not enable the model to completely suppress the background regions.

\newpara{Feature-Level Audio-Grounded Contrastive Loss.}
Suppressing masks derived from negative pairs is essential for enhancing robustness against background regions. However, due to memory constraints, generating high-resolution image-level masks for all negative pair combinations within a batch is infeasible. As an alternative, we introduce the feature-level audio-grounded contrastive loss, $ACL_F$, allowing the use of masks in lower-resolution (on features), effectively bypassing the memory constraints. A strategic approach involves emphasizing regions within the spatial visual features, as shown in Figure \ref{fig:pipeline}. To elaborate, the mask $\mathbf{M}^F_{i,j} \in \mathbb{R}^{h \times w}$ obtained from the $Masker_F$ for given image $\textbf{X}^I_i$ and audio $\textbf{X}^A_j$ is applied during spatial pooling of the spatial visual features $\boldsymbol{v}^D_i \in \mathbb{R}^{c \times h \times w}$ to focus on regions within the features that exhibit high correlation with the paired audio. Feature-level audio-grounded visual embedding $\boldsymbol{v}^F_{i,j} \in \mathbb{R}^{c}$ is as follows:
\begin{equation}
    \boldsymbol{v}^F_{i,j} = \frac{\sum_{h,w}\mathbf{M}^F_{i,j,h,w} \cdot \boldsymbol{v}^D_{i,h,w}}{\sum_{h,w}\mathbf{M}^F_{i,j,h,w}}.
        \label{eq:pooling}
\end{equation}
In contrast to $ACL_I$, which focuses on the sounding region, $ACL_F$ focuses on the highly correlated area, regardless of positive or negative audio-visual pairs. The audio-visual similarity between the audio-driven embedding $\mathbf{A}$ and the feature-level audio-grounded visual embedding $\boldsymbol{v}^F$ for both positive and negative pairs is computed using cosine similarity defined as $S^F_{i,j}=( \boldsymbol{v}^F_{i,j} \cdot \textbf{A}_j )$. The $ACL_F$ loss is defined as follows:
\begin{align}
    \mathcal L_{ACL_F} = \textit{InfoNCE}(\textbf{S}^F),
    \label{eq:aclf}
\end{align}
\noindent where $\textbf{S}^F$ is feature-level audio-visual similarity matrix within batch. While it is possible to replace the mask $\textbf{M}^F$ with $\textbf{M}^I$ in Equation~\ref{eq:pooling}, this may lead to unintended training. The reason is that $\textbf{M}^I$ may generate a mask that is close to a zero matrix when dealing with negative pairs. This can result in the numerator of Equation~\ref{eq:pooling} effectively being zero, making $\boldsymbol{v}^F_{i,j}$ arbitrary. To simplify, this replacement may cause the InfoNCE loss to generate random similarity scores for negative pairs.

\subsection{Area Regularization} \label{ssec:area}
We observe that even using $ACL_I$ and $ACL_F$ losses during training, the model can take a shortcut and output masks that contain both irrelevant and sounding regions, such as the entire image. In this case, the CLIP image encoder in the Audio-Visual Alignment module can still generate relevant visual features. Therefore, similar to~\cite{xie2022clims,cha2023learning}, we formulate an area regularizer loss, as defined below:
\begin{equation}
	\label{eq:Reg}
	\mathcal{L}_{Reg} = \sum_{i} \Vert p^+ - \overline{\mathbf{M}^I_{i,i}} \Vert_1 + \sum_{i \ne j} \Vert p^- - \overline{\mathbf{M}^I_{i,j}} \Vert_1,
\end{equation}

\noindent where $\mathbf{M}^I_{i,i}$ and $\mathbf{M}^I_{i,j}$ are the image masks from the positive and the negative pairs respectively. The area of these masks are denoted as $\overline{\mathbf{M}}$. $p^+$ and $p^-$ represent the area prior hyperparameters, which are set to 0.4 and 0.0. The area regularizer constrains the size of the mask during learning to ensure that the intended sounding regions are contained while irrelevant areas are discarded. 

\subsection{Training} \label{ssec:training}
The overall training loss term is defined as follows:
\begin{equation}
	\label{eq:final}
	\mathcal{L} = \lambda_{ACL_I} \mathcal{L}_{ACL_I} + \lambda_{ACL_F} \mathcal{L}_{ACL_F} + \lambda_{REG} \mathcal{L}_{REG},
\end{equation}
where $\lambda_{ACL_I}$, $\lambda_{ACL_F}$, and $\lambda_{REG}$ are the hyper-parameters weighting the loss terms.

\subsection{Inference} \label{ssec:inference}
For the provided image and audio pairs, an audio-driven embedding is acquired and fed into the grounder $G$ along with the visual features obtained from the image encoder. The resulting output of the grounder, $\mathbf{M}^G$, is subsequently used in $Masker_I$. Unlike training, during inference, it is adjusted using $\sigma\left(\mathbf{M}^G+ {b}/{w}\right)$, where $w$, $b$ are scalar projection parameters learned during training in the image masker $Masker_I$ and $\sigma$ is sigmoid function. The final output mask is then thresholded using the hyperparameter $t$ to obtain the localization result.

\section{Experiments}
\newpara{Datasets.} 
Our approach is trained using the VGGSound dataset~\cite{VGGSound}, comprising around \app 200K videos. After training, we evaluate sound localization performance on VGG-SS~\cite{chen2021localizing} and SoundNet-Flickr-Test~\cite{senocak2018learning,senocak2019learning} datasets. These evaluation sets provide bounding box annotations for sound sources, totaling about 5K and 250 samples, respectively. Further evaluations are conducted using AVSBench~\cite{zhou2022avs} and Extended VGG-SS/SoundNet-Flickr~\cite{slavc} datasets. AVSBench includes binary segmentation maps indicating audio-visually related pixels and is divided into Single-source (S4) and Multi-sources (MS3) subsets, categorized by the number of sounding objects. 
These subsets contain around 5K samples in (S4) and about 400 samples in (MS3). Lastly, the Extended VGG-SS/SoundNet-Flickr datasets proposed by~\cite{slavc} are used to explore non-visible sound sources.

\newpara{Implementation details.} We employ frozen pre-trained ``ViT-B/16'' CLIP~\cite{radford2021learning} model as image encoder, BEATs~\cite{chen2022beats} for audio encoder and CLIPSeg~\cite{luddecke2022image} for grounder. During training, we used 10-second audio segments sampled at 16kHz, and the center frame of the video resized to 352x352. For the overall loss, we set the parameters $\lambda_{ACL_I}$, $\lambda_{ACL_F}$, and $\lambda_{Reg}$ all to 1. Additionally, we used $\tau$ as 0.07 in Equation \ref{eq:acli}. The model is optimized for 20 epochs with a batch size of 16, using the Adam optimizer with a learning rate of $10^{-3}$ and a weight decay of $10^{-3}$.

\subsection{Quantitative Results} \label{ssec:quan}
\vspace{-2mm}
\newpara{Baselines.} Besides the existing works, we also compare our proposed method with closely-related baselines that can be obtained using different components of our overall architecture. The details of these baselines are introduced below:
\begin{itemize}
    \item \textbf{CLIPSeg w/ GT Text.} 
    We utilize the ground truth class labels of test samples as text conditions to obtain the segmentation results from CLIPSeg, essentially serves as an oracle method.
    \item \textbf{CLIPSeg w/ WAV2CLIP Text.} WAV2CLIP aligns text, vision, and audio embeddings together in the CLIP space. For a given audio, the most relevant text (class label) can be retrieved. This retrieved text is used with CLIPSeg to highlight the sounding region in the image.
    \item \textbf{CLIPSeg - Sup. AudioTokenizer} We train AudioTokenizer module in a supervised manner rather than in a self-supervised way like our proposed model. The predicted audio-driven embedding is supervised using the corresponding GT text $\textbf{X}^T $ of each sample directly with $\mathcal L_{Sup} = \lVert E_{\text{CLIP}_T}(\textbf{X}^T) - \textbf{\textbf{A}}\rVert_{1}$.
    The audio-driven embeddings obtained from this model are used with zero-shot CLIPSeg to obtain sound localization results. 
    \item \textbf{WAV2CLIP and AudioCLIP.} These models leverage the pre-trained CLIP model to align text, vision, and audio embeddings. To enable zero-shot sound source localization with these models, we utilize a pre-trained CLIP-like object detector~\cite{li2022glip} to extract region proposals from the images and calculate the cosine similarity between the visual features of those regions and the audio features. The region with the highest similarity is employed as the localization result.
\end{itemize}

\begin{table}
    \centering
    \resizebox{1.0\linewidth}{!}{
    \begin{tabular}{lccccc}
    \toprule
    & \multicolumn{2}{c}{\textbf{VGG-SS}} & \multicolumn{2}{c}{\textbf{SoundNet-Flickr}} \\
    \textbf{Method} &  \textbf{cIoU $\uparrow$} & \textbf{AUC $\uparrow$} & \textbf{cIoU $\uparrow$} & \textbf{AUC $\uparrow$} \\ \midrule
    Attention~\cite{senocak2018learning}$_{\text{CVPR}18}$ &  18.50 & 30.20 & 66.00 & 55.80 \\
    CoarseToFine~\cite{qian2020multiple}$_{\text{ECCV}20}$ & 29.10 & 34.80 & - & - \\
    LCBM~\cite{senocakLessMore}$_{\text{WACV}22}$ & 32.20 & 36.60 & - & - \\
    LVS~\cite{chen2021localizing}$_{\text{CVPR}21}$ & 34.40 & 38.20 & 71.90 & 58.20 \\
    HardPos~\cite{senocakHardPos}$_{\text{ICASSP}22}$ & 34.60 & 38.00 & 76.80 & 59.20 \\
    SSPL~\cite{song2022sspl}$_{\text{CVPR}22}$ & 33.90 & 38.00 & 76.70 & 60.50 \\
    EZ-VSL (w/o OGL)~\cite{ezvsl}$_{\text{ECCV}22}$ & 35.96 & 38.20 & 78.31 & 61.74 \\
    EZ-VSL (w/ OGL)~\cite{ezvsl}$_{\text{ECCV}22}$ & 38.85 & 39.54 & 83.94 & 63.60 \\
    SSL-TIE~\cite{ssslTransformation}$_{\text{ACM MM}22}$ & 38.63 & 39.65 & 79.50 & 61.20 \\
    SLAVC (w/o OGL)~\cite{slavc}$_{\text{NeurIPS}22}$ & 37.79 & 39.40 & 83.60 & - \\
    SLAVC (w/ OGL)~\cite{slavc}$_{\text{NeurIPS}22}$ & 39.80 & - & \textbf{86.00} & - \\
    MarginNCE (w/o OGL)~\cite{marginnce}$_{\text{ICASSP}23}$ & 38.25 & 39.06 & 83.94 & 63.20 \\
    MarginNCE (w/ OGL)~\cite{marginnce}$_{\text{ICASSP}23}$ & 39.78 & 40.01 & 85.14 & 64.55 \\
    HearTheFlow~\cite{htf}$_{\text{WACV}23}$ & 39.40 & 40.00 & 84.80 & 64.00 \\
    FNAC (w/o OGL)~\cite{sun2023learning}$_{\text{CVPR}23}$ & 39.50 & 39.66 & 84.73 & 63.76 \\
    FNAC (w/ OGL)~\cite{sun2023learning}$_{\text{CVPR}23}$ & 41.85 & 40.80 & 85.14 & 64.30 \\
    Alignment (w/o OGL)~\cite{senocak2023alignment}$_{\text{ICCV}23}$ & 39.94 & 40.02 & 79.60 & 63.44 \\ 
    Alignment (w/ OGL)~\cite{senocak2023alignment}$_{\text{ICCV}23}$ & 42.64 & 41.48 & 82.40 & 64.60 \\    \bottomrule
    \textit{Baselines:}   &  &  &  &  \\
    WAV2CLIP~\cite{wu2022wav2clip}$_{\text{ICASSP}22}$ & 37.71 & 39.93 & 26.00 & 29.60 \\
    AudioCLIP~\cite{guzhov2022audioclip}$_{\text{ICASSP}22}$ & 44.15 & 46.23 & 47.20 & 45.22 \\
    CLIPSeg (w/ GT Text)  & 49.50 & 48.62 & - & - \\
    CLIPSeg (w/ WAV2CLIP Text) & 24.84 & 26.01 & 37.20 & 32.14 \\
    CLIPSeg (Sup. AudioTokenizer) & 49.09 & 45.75 & 68.00 & 54.96 \\
    \rowcolor{lightgray!25}
    \textbf{Ours (w/o OGL)} & \textbf{49.46} & \textbf{46.32} & 80.80 & \textbf{64.62} \\
    \bottomrule
    \end{tabular}}
    
    \caption{\textbf{Quantitative results on the VGG-SS and SoundNet-Flickr test sets}. All models are trained with 144K samples from VGG-Sound. SLAVC~\cite{slavc} does not provide AUC scores. SoundNet-Flickr has no GT text.}\label{tab:quantitative}
    \vspace{-6mm}
\end{table}

\newpara{Comparison on standard benchmarks.} In this section, we perform a comparative analysis of our method for localizing sound sources in comparison to existing approaches and the strong baselines. Our evaluations are conducted within the established setting, similar to prior methodologies~\cite{chen2021localizing,senocakHardPos,ezvsl,sun2023learning}. We train our model on the VGGSound-144K dataset and subsequently assess its performance on the VGG-SS and SoundNet-Flickr test sets. It is worth noting that all the models we compare are trained using equivalent amounts of data. However, note that our model does not use object guided refinement (OGL). We present our findings in~\Tref{tab:quantitative}. 

At the outset, we compare our method with other existing sound source localization models. There is a substantial gap between the existing self-supervised methods and ours in VGG-SS evaluation task. Although our model is also purely trained in a self-supervised manner with the audio-visual correspondence objective, it is evident that leveraging CLIP's strong multimodal alignment knowledge significantly impacts the performance. However, note that even though we leverage CLIP, we do not employ any explicit text input. These results thus demonstrate that our AudioTokenizer module effectively encodes the audio context, enabling proper learning of the audio-visual correspondence objective. Interestingly, we observe that the zero-shot performance of our model on the SoundNet-Flickr test lags behind that of the existing models. We hypothesize that this result stems from the fact that our model generates more fine-grained outputs, resembling segmentation. Nonetheless, the ground-truth bounding boxes are relatively coarse, causing our method to yield lower cIoU scores despite successfully highlighting the sounding region. We provide some illustrative qualitative results for this in~\Sref{ssec:qual}.

Next, we conduct comparisons against the strong baselines introduced earlier. Our method outperforms or achieves on-par performance with these baselines. It is worth noting that our method does not explicitly utilize text information to highlight object regions via CLIPSeg or learn audio-driven embeddings in a supervised fashion, as done by these baselines. This indicates that our audio-visual correspondence objective effectively learns robust audio-visual correspondence and drives the AudioTokenizer and Audio-Driven Embedder to accurately project the true audio context into audio-driven embeddings. Interestingly, our model gives on-par performance with the CLIPSeg w/ GT Text baseline on VGG-SS, which serves as an Oracle. This model is text-conditioned open-world segmentation approach and utilizes the ground-truth class labels of the test samples. 
This signifies that it is important to incorporate the audio context properly to enhance performance. Additionally, the performance difference between CLIPSeg w/ GT Text and CLIPSeg w/ WAV2CLIP highlights that the text-queried zero-shot performance of CLIPSeg in sound source localization is highly dependent on the quality of the text input. This is due to the fact that the text retrieved from WAV2CLIP for given audio tends to be noisier compared to GT text. Nevertheless, it is important to note that sound source localization is unlabeled task, and these methods serve as Oracle baselines. Furthermore, the results demonstrate that training the AudioTokenizer in a supervised way with GT texts and employing audio-driven embeddings with CLIPSeg also gives on-par performance to the Oracle. This implies that audio-driven embeddings indeed provide accurate information for highlighting the sounding regions. Finally, we acknowledge that our method, which employs an audio-visual correspondence objective for self-supervised learning, outperforms CLIPSeg - Sup. AudioTokenizer. This suggests that our Audio-Visual alignment contrastive losses offer effective supervision for the model as using explicit text input, compelling the AudioTokenizer module to generate richer audio-driven embeddings.

\begin{table}
    \centering
    \footnotesize
    \scalebox{0.75}{
    \begin{tabular}{l|lcc}
    \toprule
    \textbf{Test Class} & \textbf{Method} & \textbf{cIoU $\uparrow$} & \textbf{AUC $\uparrow$} \\ 
    \hline 
    \multirow{9}{*}{Heard 110}    
    & LVS~\cite{chen2021localizing}$_{\text{CVPR}21}$ & 28.90 & 36.20 \\
    & EZ-VSL (w/o OGL)~\cite{ezvsl}$_{\text{ECCV}22}$ & 31.86 & 36.19 \\
    & EZ-VSL (w/ OGL)~\cite{ezvsl}$_{\text{ECCV}22}$ & 37.25 & 38.97 \\
    & SLAVC (w/o OGL)~\cite{slavc}$_{\text{NeurIPS}22}$ & 35.84 & - \\
    & SLAVC (w/ OGL)~\cite{slavc}$_{\text{NeurIPS}22}$ & 38.22 & - \\
    & FNAC (w/ OGL)~\cite{sun2023learning}$_{\text{CVPR}23}$ & 39.54 & 39.83 \\
    & Alignment (w/o OGL)~\cite{senocak2023alignment}$_{\text{ICCV}23}$ & 38.31 & 39.05 \\
    & Alignment (w OGL)~\cite{senocak2023alignment}$_{\text{ICCV}23}$ & 41.85 & 40.93 \\
    \cline{2-4}
    & CLIPSeg (w/ GT Text) & 49.65 & 45.74 \\
    & CLIPSeg (w/ WAV2CLIP Text) & 23.24 & 24.78 \\
    & CLIPSeg (Sup. AudioTokenizer) & 49.73 & 45.35 \\
    & \cellcolor{lightgray!25}\textbf{Ours (w/o OGL)} & \cellcolor{lightgray!25} 48.44 & \cellcolor{lightgray!25} 45.06 \\
    \hline \hline
    
    \multirow{9}{*}{Unheard 110} 
    & LVS~\cite{chen2021localizing}$_{\text{CVPR}21}$ & 26.30 & 34.70 \\
    & EZ-VSL (w/o OGL)~\cite{ezvsl}$_{\text{ECCV}22}$ & 32.66 & 36.72 \\
    & EZ-VSL (w/ OGL)~\cite{ezvsl}$_{\text{ECCV}22}$ & 39.57 & 39.60 \\
    & SLAVC (w/o OGL)~\cite{slavc}$_{\text{NeurIPS}22}$ & 36.50 & - \\
    & SLAVC (w/ OGL)~\cite{slavc}$_{\text{NeurIPS}22}$ & 38.87 & - \\
    & FNAC (w/ OGL)~\cite{sun2023learning}$_{\text{CVPR}23}$ & 42.91 & 41.17 \\
    & Alignment (w/o OGL)~\cite{senocak2023alignment}$_{\text{ICCV}23}$ & 39.11 & 39.80 \\
    & Alignment (w OGL)~\cite{senocak2023alignment}$_{\text{ICCV}23}$ & 42.94 & 41.54 \\
    \cline{2-4}
    & CLIPSeg (w/ GT Text) & 49.13 & 44.77 \\
    & CLIPSeg (w/ WAV2CLIP Text) & 26.25 & 27.03 \\
    & CLIPSeg (Sup. AudioTokenizer) & 43.65 & 41.05 \\
    & \cellcolor{lightgray!25}\textbf{Ours (w/o OGL)} & \cellcolor{lightgray!25} 41.98 & \cellcolor{lightgray!25} 41.55 \\
    \bottomrule
    \end{tabular}}
    \caption{\textbf{Comparison results on open-set audio-visual localization experiments trained and tested on the splits of~\cite{chen2021localizing,ezvsl,marginnce}.}} \label{tab:exp_openset_first}
    \vspace{-6mm}
\end{table}

Finally, we compare our model with AudioCLIP and WAV2CLIP, both of which are contrastively trained on image-audio pairs, leveraging the pre-trained CLIP. The results in~\Tref{tab:quantitative} demonstrate that our method outperforms these approaches. This indicates that our Audio-Driven Embedder module, with the audio-visual alignment objective, is more effective in learning a stronger audio-visual alignment than these previous approaches, as they also leverage pre-trained CLIP knowledge. Additionally, note that these baseline approaches incorporate powerful object detectors to obtain object proposals/areas that correspond with the given audio, in order to achieve sound localization results.

\newpara{Open Set Audio-Visual Localization.} Chen et al.~\cite{chen2021localizing} propose an open-set evaluation scenario to assess the generalization ability of sound source localization methods. This evaluation setting involves testing the models on categories present in the training data (heard), as well as categories that are absent (unheard). For this evaluation, 110 randomly selected categories from the VGGSound dataset are used for training, while an entirely separate set of 110 categories is held for testing. This ensures that the model encounters new and previously unseen categories during the evaluation process. To make a fair comparison, we conduct the experiments using the the same train/test split as~\cite{chen2021localizing,ezvsl,marginnce}. It is important to note that unlike previous methods, we do not utilize object-guided refinement (OGL). The results are presented in~\Tref{tab:exp_openset_first}, showing that our method outperforms existing approaches in the Heard categories. However, it lags behind FNAC~\cite{sun2023learning} in the Unheard category, due to the usage of OGL in their method, which we do not employ.

\newpara{Extended Flickr-SoundNet/VGG-SS.} Existing benchmarks typically consist of sounding objects/regions in the scene. However, in reality, silent objects or off-screen audio are also common occurrences. Mo et al.~\cite{slavc} propose a new evaluation that extends the existing benchmarks to include non-audible frames, non-visible sound sources, and mismatched audio-visual pairs. In this evaluation scenario, it is expected that sound localization methods should not highlight an object/region if the audio and visual signals are mismatched. The experiments conducted using the extended Flickr-SoundNet/VGG-SS datasets in~\Tref{tab:extended} demonstrate that our method outperforms all the existing methods and baselines. The superiority of our method indicates that it learns a strong alignment of audio and visual embeddings with the help of our AudioTokenizer and leveraging CLIP without text input, as this task requires a robust semantic relationship between the cross-modalities. One interesting observation is that, even though baseline approaches leverage CLIP, their performance is lower than ours due to the absence of audio-visual alignment supervision. 

\begin{table}
    \centering
    \resizebox{1.0\linewidth}{!}{
    \begin{tabular}{lcccccc}
    \toprule
     & \multicolumn{3}{c}{\textbf{Extended VGG-SS}} & \multicolumn{3}{c}{\textbf{Extended Flickr}} \\
    Method & \textbf{AP $\uparrow$} & \textbf{max-F1 $\uparrow$} & \textbf{LocAcc $\uparrow$} & \textbf{AP $\uparrow$} & \textbf{max-F1 $\uparrow$} & \textbf{LocAcc $\uparrow$} \\ \midrule
    SLAVC (w/o OGL)~\cite{slavc}$_{\text{NeurIPS}22}$ & 32.95 & 40.00 & 37.79 & 51.63 & 59.10 & 83.60 \\
    MarginNCE (w/o OGL)~\cite{marginnce}$_{\text{ICASSP}23}$ & 30.58 & 36.80 & 38.25 & 57.99 & 61.80 & 83.94\\
    FNAC (w/o OGL)~\cite{sun2023learning}$_{\text{CVPR}23}$ & 23.48 & 33.70 & 39.50 & 50.40 & 62.30 & \textbf{84.73} \\
    Alignment (w/o OGL)~\cite{senocak2023alignment}$_{\text{ICCV}23}$ & 34.73 & 40.70 & 39.94 & 64.43 & 66.90 & 79.60 \\

    WAV2CLIP~\cite{wu2022wav2clip}$_{\text{ICASSP}22}$ & 26.67 & 33.00 & 37.71 & 20.99 & 24.80 & 29.60 \\
    AudioCLIP~\cite{guzhov2022audioclip}$_{\text{ICASSP}22}$ & 23.79 & 32.80 & 44.15 & 34.00 & 38.80 & 45.22 \\
    CLIPSeg (Sup. AudioTokenizer) & 34.96 & 41.00 & 49.09 & 55.14 & 57.00 & 68.00 \\    
    \rowcolor{lightgray!25}
    \textbf{Ours (w/o OGL)}  & \textbf{40.79} & \textbf{49.10} & \textbf{49.46} & \textbf{76.07} & \textbf{73.20} & 80.80 \\    
    \bottomrule
    \end{tabular}}
    \vspace{-2mm}
    \caption{\textbf{Quantitative results on Extended VGG-SS and Extended Flickr-SoundNet benchmark.} All models are trained with 144K samples from VGG-Sound. The results of the prior approaches are obtained from~\cite{slavc}.}
    \label{tab:extended}
    \vspace{-6mm}
\end{table}

\begin{figure*}[tp]
    \centering
    \includegraphics[width=\linewidth]{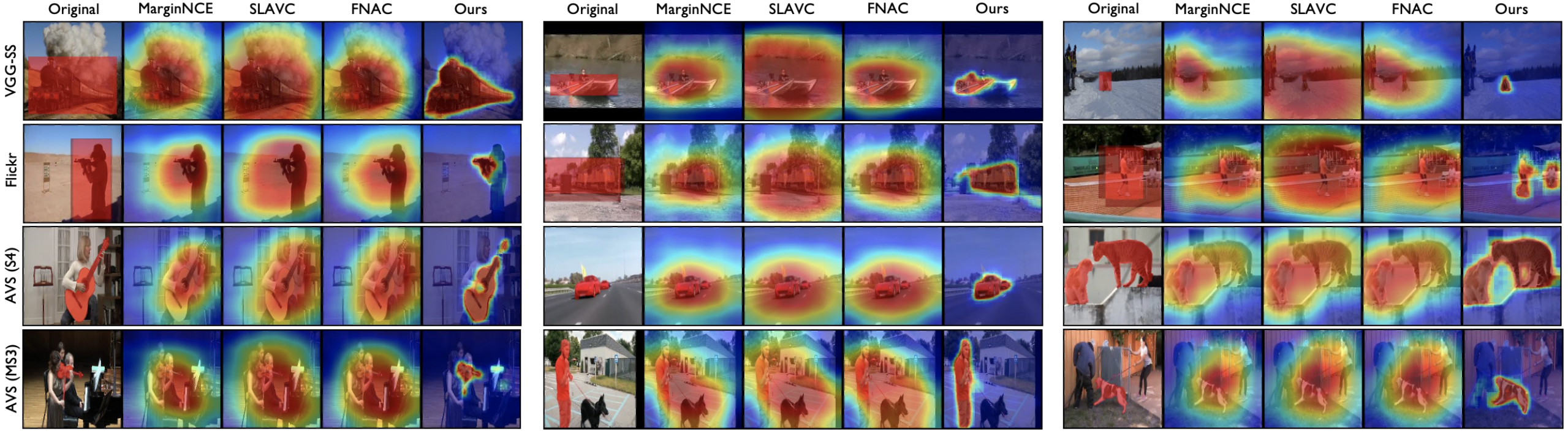}
    \caption{{\bf 
    Sound localization results on VGG-SS, SoundNet-Flickr, and AVSBench datasets, along with a comparison with previous methods.
    } }
    \label{fig:qualitatives}
\vspace{-6mm}
\end{figure*}

\newpara{AVSBench~\cite{zhou2022avs}.} We conduct additional experiments using the AVSBench S4 and MS3 datasets to demonstrate the precise sound localization ability of our model. These datasets are designed to identify audio-visual correspondences at the pixel level, \ie audio-visual segmentation. In these experiments, all models are trained on VGGSound-144K and then tested on the AVSBench datasets in a zero-shot setting. Our results, presented in~\Tref{tab:avs}, draw a substantial performance gap compared to existing methods. This gap is more pronounced on audio-visual segmentation datasets than on standard benchmarks, as our model tends to generate more fine-grained localization maps due to the grounder and learnable maskers it employs. Our proposed method also demonstrates competitive or stronger performance compared to most of the baselines. While our method outperforms others on the S4 dataset, CLIPSeg w/ GT Text and CLIPSeg w/ WAV2CLIP Text (Oracles) achieve better segmentation performance on the MS3 dataset. However, we emphasize that our model does not employ any direct supervision or usage of the text. Instead, it relies solely on audio-visual alignment. Also, note that sound source localization task is theoretically unlabeled. 

\begin{table}
    \centering
    \resizebox{1.0\linewidth}{!}{
    \begin{tabular}{lccccc}
    \toprule
    & \multicolumn{2}{c}{\textbf{S4}} & \multicolumn{2}{c}{\textbf{MS3}} \\
    \textbf{Method} &  \textbf{mIoU $\uparrow$} & \textbf{F-Score $\uparrow$} & \textbf{mIoU $\uparrow$} & \textbf{F-Score $\uparrow$} \\ \midrule
    SLAVC (w/o OGL)~\cite{slavc}$_{\text{NeurIPS}22}$ & 28.10 & 34.60 & 24.37 & 25.56 \\
    MarginNCE (w/o OGL)~\cite{marginnce}$_{\text{ICASSP}23}$ & 33.27 & 45.33 & 27.31 & 31.56 \\
    FNAC (w/o OGL)~\cite{sun2023learning}$_{\text{CVPR}23}$ & 27.15 & 31.40 & 21.98 & 22.50 \\
    Alignment (w/o OGL)~\cite{senocak2023alignment}$_{\text{ICCV}23}$ & 29.60 & 35.90 & - & - \\

    \bottomrule
    \textit{Baselines:}   &  &  &  &  \\
    WAV2CLIP~\cite{wu2022wav2clip}$_{\text{ICASSP}22}$ & 28.70 & 35.35 & 25.09 & 23.84 \\
    AudioCLIP~\cite{guzhov2022audioclip}$_{\text{ICASSP}22}$ & 36.57 & 42.15 & 27.06 & 26.48 \\
    CLIPSeg (w/ GT Text)  & 51.32 & 58.02 & \textbf{50.93} & \textbf{55.41} \\
    CLIPSeg (w/ WAV2CLIP Text) & 26.52 & 30.60 & 30.82 & 29.97 \\
    CLIPSeg (Sup. AudioTokenizer) & 49.82 & 56.43 & 42.57 & 46.72 \\
    \rowcolor{lightgray!25}
    \textbf{Ours (w/o OGL)} & \textbf{59.76} & \textbf{69.03} & 41.08 & 46.67 \\
    \bottomrule
    \end{tabular}}
    \vspace{-2mm}
    \caption{\textbf{Quantitative results on the AVSBench test sets}.}\label{tab:avs}
    \vspace{-4mm}
\end{table}

\begin{table}
    \centering
    \resizebox{1.0\linewidth}{!}{
    \begin{tabular}{lccccccccc}
    \toprule
    &\multicolumn{3}{c}{} & \multicolumn{2}{c}{\textbf{VGG-SS}} & \multicolumn{2}{c}{\textbf{AVS (S4)}} & \multicolumn{2}{c}{\textbf{Extended VGG-SS}} \\
    \textbf{} & $ACL_I$ & $ACL_F$ & $Reg$ & \textbf{cIoU $\uparrow$} & \textbf{AUC $\uparrow$} & \textbf{mIoU $\uparrow$} & \textbf{F-score $\uparrow$} & \textbf{AP $\uparrow$} & \textbf{max-F1 $\uparrow$} \\ \midrule
    (A)  & \ding{51} & \ding{55} & \ding{55} & 40.42 & 40.84 & 38.55 & 45.94 & 28.59 & 35.90  \\
    (B)  &  \ding{55} & \ding{51} & \ding{55} & 2.30 & 7.46 & 4.08 & 22.59 & 0.86 & 1.80  \\
    (C)  & \ding{51} & \ding{51} & \ding{55} & 46.61 & 44.71 & 53.06 & 63.01 & 40.72 & 47.90  \\
    (D)  & \ding{51} & \ding{55} & \ding{51} & 41.08 & 41.01 & 41.93 & 48.99 & 33.37 & 41.30  \\
    (E)  & \ding{55} & \ding{51} & \ding{51} & 35.15 & 38.36 & 32.06 & 41.05 & 39.91 & 47.20  \\
    \rowcolor{lightgray!25}
    (F)  & \ding{51} & \ding{51} & \ding{51} & \textbf{49.46} & \textbf{46.32} & \textbf{59.76} & \textbf{69.03} & \textbf{40.79} & \textbf{49.10}  \\    
    \bottomrule
    \end{tabular}}
    \vspace{-2mm}
    \caption{\textbf{Ablative experiments on our method by using different combinations of loss functions.}}
    \label{tab:ablation}
    \vspace{-6mm}
\end{table}

\vspace{-2mm}
\subsection{Ablation Results} \label{ssec:ablation} 
Our proposed method is optimized by a combination of three loss functions, \ie $ACL_I$, $ACL_F$, and area regularization. Here, we perform ablation experiments to understand the impact of each loss function. We primarily conduct the experiments by training our model on VGGSound-144K and evaluating it on VGG-SS, AVSBench and Extended VGG-SS datasets. Results are in~\Tref{tab:ablation}.

As revealed by results (A) and (B), using $ACL_I$ is crucial to enable our model to learn the corresponding audio-visual alignment. On the other hand, relying solely on $ACL_F$ is not effective for learning audio-visual alignment, as it primarily focuses on suppressing unrelated areas. However, as demonstrated by the results of (A \emph{vs.} C) and (B \emph{vs.} C), the combination of these two loss functions are complementary. As mentioned earlier, $ACL_F$ contributes to performance enhancement by suppressing background areas. Furthermore, an examination of the results from the experiments (C \emph{vs.} F) highlights that area regularization provides additional improvements by constraining the size of the activated regions. Visualization of these ablative studies can be found in~\Fref{fig:qualitative_ablation}.

\subsection{Qualitative Results} \label{ssec:qual}

\vspace{-2mm}
\newpara{Comparison to the existing approaches.} ~\Fref{fig:qualitatives} displays the comparison results between our method and recent prior works. The visualized samples illustrate that the localized regions from our proposed method are more compact and fine-grained compared to the other methods. For example, regardless of the test set, our model can accurately localize small-sized sounding objects compared to recent methods. Moreover, our model accurately highlights multiple sound sources and separates them, while other methods tend to cover the entire area as one large region (last column of the second and third rows).

\begin{figure}[tp]
    \centering
    \includegraphics[width=\linewidth]{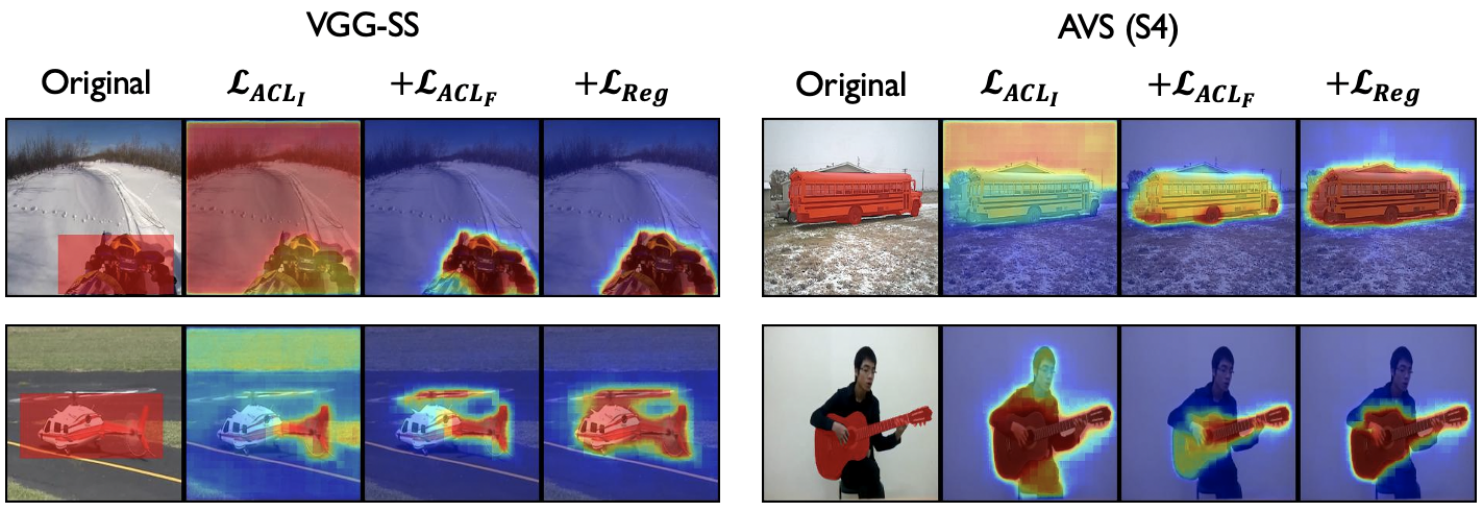}
    \vspace{-4mm}
    \caption{{\bf Sound localization results by using different combinations of loss functions.} }
    \label{fig:qualitative_ablation}
\vspace{-4mm}
\end{figure}

\begin{figure}[tp]
    \centering
    \includegraphics[width=\linewidth]{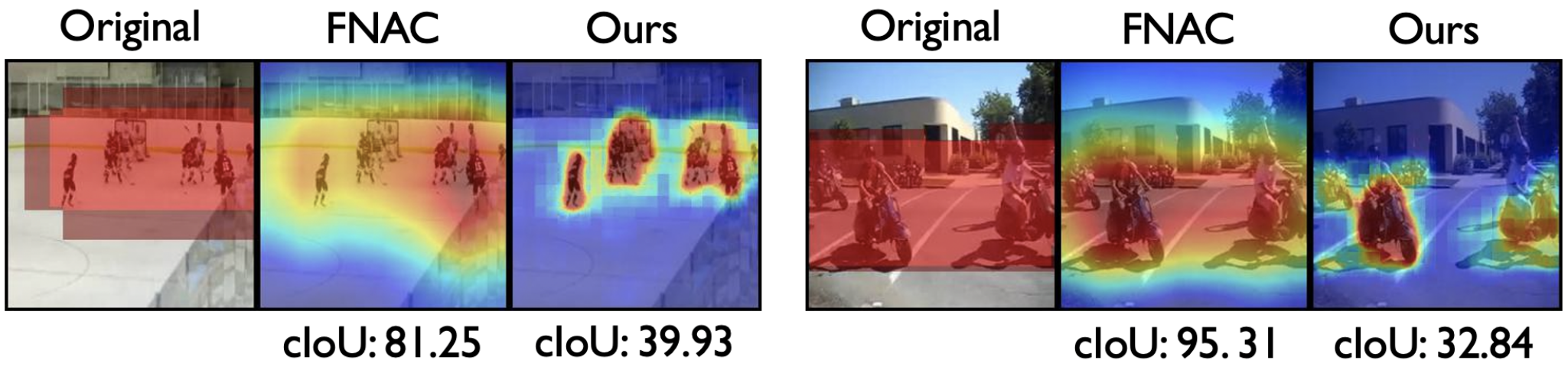}
    \vspace{-6mm}
    \caption{{\bf cIoU scores of SoundNet-Flickr samples.} }
    \label{fig:qualitative_cIoU}
\vspace{-6mm}
\end{figure}

\newpara{Visualization of the ablation experiments.} The visual results are presented in~\Fref{fig:qualitative_ablation}. As demonstrated, when using only $ACL_I$, we observe that background areas remain activated (also discussed in~\Sref{ssec:alignment}). As evident in the third column, the addition of $ACL_F$ helps eliminate the background pixels (non-sounding areas). However, it is noticeable that the outputs of $ACL_I$+$ACL_F$ can be relatively less completed. With the area regularizer, the final output of our model becomes more complete and fine-grained.

\newpara{Visualization of fine-grained localization with lower cIoU.} We present our localization results along with the cIoU scores on SoundNet-Flickr. As depicted in~\Fref{fig:qualitative_cIoU}, despite our model successfully highlighting the sounding regions, these results yield lower cIoU scores. This outcome is consistent with the quantitative results in~\Tref{tab:quantitative}, which demonstrate that our method on SoundNet-Flickr lags behind the other methods due to the fact that the GT boxes and the localization results of competing methods are coarse.

\section{Conclusion}
In this work, we explore using large-scale pre-trained image-text models, specifically CLIP, for sound source localization. Our aim is to integrate CLIP's multimodal alignment knowledge in a text input-free form through self-supervised audio-visual correspondence. To this end, we translate audio signals into CLIP-compatible tokens and use the resulting audio-driven embeddings for audio-visual grounding. This process is integrated with contrastive learning, enabling self-supervised audio-visual alignment learning. We show that our proposed model significantly outperforms existing methods in audio-visual coarse sound source localization and fine-grained segmentation tasks. Moreover, it compares favorably with fully supervised or text-queried baselines. Our study suggests that the true essence of sound source localization, characterized by strong audio-visual alignment, can take advantage from the already structured multimodal alignment offered by large-scale pre-trained image-text models.

{
\clearpage
\small
\bibliographystyle{ieee_fullname}
\bibliography{egbib}
}

\end{document}